%% file: paper_p1_arxiv.tex
\input{paper_p1_arxiv_metadata.tex}
\ifdefined\PoneArxiv
\documentclass[sigconf,nonacm]{acmart}
\else
\documentclass[sigconf,anonymous,review]{acmart}
\fi

\usepackage{amsmath,mathtools}
\usepackage{booktabs}
\usepackage{array}
\usepackage{algorithm}
\usepackage{algpseudocode}
\counterwithin{algorithm}{section}
\usepackage{tikz}
\usetikzlibrary{arrows.meta,positioning,fit,calc,decorations.pathreplacing}
\usepackage[T1]{fontenc}
\usepackage[utf8]{inputenc}
\DeclareUnicodeCharacter{2192}{\ensuremath{\to}}
\DeclareUnicodeCharacter{2190}{\ensuremath{\leftarrow}}
\DeclareUnicodeCharacter{2264}{\ensuremath{\le}}
\DeclareUnicodeCharacter{2248}{\ensuremath{\approx}}
\DeclareUnicodeCharacter{221E}{\ensuremath{\infty}}
\DeclareUnicodeCharacter{2113}{\ensuremath{\ell}}
\DeclareUnicodeCharacter{03B1}{\ensuremath{\alpha}}
\DeclareUnicodeCharacter{0394}{\ensuremath{\Delta}}
\DeclareUnicodeCharacter{03C3}{\ensuremath{\sigma}}
\DeclareUnicodeCharacter{03A6}{\ensuremath{\Phi}}
\DeclareUnicodeCharacter{2010}{-}
\DeclareUnicodeCharacter{2011}{-}
\DeclareUnicodeCharacter{2013}{--}
\DeclareUnicodeCharacter{2014}{---}

\newcommand{\R}{\mathbb{R}}

\ifdefined\PoneArxiv
\PoneArxivPdfMetadata
\fi

\begin{document}

\title{Cryptanalytic Extraction of Isolated Bias-Free GLU Feed-Forward Blocks by Antipodal Separation}

\ifdefined\PoneArxiv
\PoneArxivAuthors
\else
\author{Anonymous Author(s)}
\affiliation{%
  \institution{Anonymous Institution}
  \country{}
}

\acmConference[AISec '26]{19th ACM Workshop on Artificial Intelligence and Security}{November 2026}{The Hague, Netherlands}
\acmYear{2026}
\copyrightyear{2026}
\setcopyright{acmlicensed}
\fi

\begin{abstract}
Cryptanalytic extraction has been demonstrated for ReLU networks, for networks using componentwise activations such as GELU or SiLU, and for a Transformer's final projection matrix. These methods do not recover the bias-free Gated Linear Unit (GLU) feed-forward blocks used in many modern language models. Such a block multiplies an activated linear projection by a second learned linear projection within each hidden unit, a two-branch structure absent from the network classes and final-layer setting addressed by those methods.

We give a constructive, multi-stage forward-query recovery primitive for isolated bias-free GLU blocks. Finite-difference curvature supplies gate-direction candidates, and paired observations at \(x\) and \(-x\) separate gate magnitude, orientation, and value-branch coupling. Across high-precision targets, six Qwen layers, an 8,192-unit Llama subproblem, and a full-dimensional Gemma block all reach sub-percent median validation error. Four finite-precision configurations remain below 5\% median error, but none reproduces every stored weight. These isolated-block experiments are not an end-to-end model-API attack: deriving the required internal block responses from final model outputs remains unsolved.
\end{abstract}



\maketitle

\section{Introduction}\label{introduction}
Exact model extraction asks whether private weights can be recovered by choosing inputs and observing outputs. Existing cryptanalytic attacks recover fully connected ReLU networks~\cite{CJM20,CanalesMartinez24} and non-gated chains using componentwise GELU, SiLU, and related activations~\cite{ADFM26}. However, these methods do not address gated units that multiply two learned branches. Dauphin et al. introduced this architecture as the Gated Linear Unit (GLU), and Shazeer later adapted it into Transformer feed-forward network (FFN) variants such as GeGLU and SwiGLU~\cite{Dauphin17,Shazeer20}. The original GLU and Shazeer's general variants include branch biases, whereas Shazeer omits them in the Transformer variants evaluated in that work. Every checkpoint architecture evaluated here uses this bias-free form. We therefore study it in isolation:
\[
y(x) = W_o \big(\sigma(W_g x) \odot (W_u x)\big),
\]
where \(W_g,W_u\) map the input \(x\) to the hidden dimension, \(W_o\) maps the product to the output, and \(\odot\) is coordinatewise multiplication. Nonzero gate, value, or output biases are outside our scope. We test SiLU-gated SwiGLU and GELU-gated GeGLU.

\subsection{Experimental Query Setting}\label{experimental-query-setting}
We use the forward-query threat model of Carlini, Jagielski, and Mironov~\cite{CJM20}: the attacker chooses each input to the target and observes the corresponding output, but receives no gradients, backpropagation access, or internal activations. Our experimental target is a single isolated GLU FFN \(f:\mathbb{R}^n\to\mathbb{R}^{n_{\mathrm{out}}}\). For each query, the attacker supplies \(x\in\mathbb{R}^n\) and observes \(y(x)=f(x)\in\mathbb{R}^{n_{\mathrm{out}}}\).

The recovery procedure is given the block dimensions \(n,m,n_{\mathrm{out}}\), the gate activation and its implementation, the returned-output format, and the declared storage format used only by Stage~3. These are architecture or interface metadata, not recovered weights. The target weights and intermediate values remain secret, and checkpoint values enter only evaluation after a recovery is fixed. We do not evaluate recovery under misspecified metadata.

In a complete Transformer, tokenization and preceding computations produce the FFN input \(x\); residual composition, later blocks, and the language-model head transform \(f(x)\) into the API output. We directly query only \(x\mapsto f(x)\) and neither construct \(x\) nor infer \(f(x)\) from model outputs (Figure~\ref{fig:scope}).

\input{figures/scope}

Because the recovery uses finite differences of target outputs, target arithmetic is also part of the experimental setting. Prior cryptanalytic extraction experiments commonly use high-precision targets~\cite{CJM20,CanalesMartinez24,FMSH24,ADFM26,QLWSW26}. We evaluate 64-bit floating-point (FP64), bfloat16 (BF16), half-precision (FP16), single-precision (FP32), and FP32-stored TensorFloat-32 (TF32) target paths; Section~\ref{empirical-evaluation} specifies each backend.

\subsection{GLU Recovery Strategy}\label{glu-recovery-strategy}
Prior attacks isolate a unit through one scalar activation's local behavior~\cite{CJM20,ADFM26}. A GLU unit instead couples one row from each branch with one output column. We separate them using two query-derived observations. Symmetric finite differences estimate \(H_k(x)=\nabla_x^2 f_k(x)\), the matrix of second input derivatives of output coordinate \(f_k\); each estimate is a \emph{Hessian observation}. We call paired responses at \(x\) and \(-x\) \emph{antipodal}.

Let \(w_{g,\ell},w_{u,\ell}\in\R^n\) denote row \(\ell\) of \(W_g,W_u\), and let \(W_o[:,\ell]\) denote column \(\ell\) of \(W_o\). The contribution of hidden unit \(\ell\) to the block output is
\[
f_\ell(x)=W_o[:,\ell]\,\sigma(w_{g,\ell}^\top x)(w_{u,\ell}^\top x),
\qquad
f(x)=\sum_{\ell=1}^{m}f_\ell(x).
\]
This unit-wise coupling motivates four quantities that the recovery separates:

\begin{itemize}
\item \textbf{gate direction} \(v_\ell\), an arbitrary unit representative of the unoriented line spanned by \(w_{g,\ell}\);
\item \textbf{gate magnitude} \(c_\ell = \|w_{g,\ell}\| > 0\), one positive scalar;
\item \textbf{gate orientation} \(s_\ell \in \{\pm 1\}\), the sign of \(w_{g,\ell}\) relative to \(v_\ell\), which Stage~1 cannot fix;
\item \textbf{coupling} \(C_\ell\), the rank-1 matrix \(W_o[:,\ell]w_{u,\ell}^\top\) through which the value branch enters the map.
\end{itemize}

The gate row is \(w_{g,\ell} = s_\ell c_\ell v_\ell\), and \((w_{u,\ell}^\top x)W_o[:,\ell] = C_\ell x\). The factorization of \(C_\ell\) into \(w_{u,\ell}\) and \(W_o[:,\ell]\) has a scalar ambiguity.

Two facts enable the separation. First, each hidden unit contributes to the Hessian a term proportional to the rank-1 matrix \(w_{g,\ell}w_{g,\ell}^\top\), which motivates a search for gate-direction candidates in a shared Hessian subspace. Second, SiLU is affine-symmetric in the terminology of Asselineau et al.~\cite{ADFM26}: it decomposes into odd and even scalar parts,
\[
\mathrm{silu}(z)
= \underbrace{\tfrac{z}{2}}_{\text{odd}}
+ \underbrace{\tfrac{z}{2}\tanh\!\tfrac{z}{2}}_{\text{even}}.
\]
For the vector output, define
\[
O(x)=\tfrac12\bigl(f(x)-f(-x)\bigr),
\qquad
E(x)=\tfrac12\bigl(f(x)+f(-x)\bigr).
\]
We call these parts odd and even because \(O(-x)=-O(x)\) and \(E(-x)=E(x)\), coordinatewise. Asselineau et al. use affine symmetry to analyze sign ambiguity in componentwise chains; our contribution is to exploit the parallel GLU value branch so that \(O(x)\) is unchanged by gate orientation and \(E(x)\) depends linearly on it. The Hessian view supplies direction candidates, \(O(x)\) fits magnitudes and couplings, and \(E(x)\) supports the orientation solve. Section~\ref{the-attack-separation-by-observation} derives these relations.

The pipeline forms and repairs Hessian-based candidates, fits magnitudes and couplings, estimates orientations, refines the continuous weights, and selects a representative in the declared storage format. We evaluate its result at two levels: \emph{functional recovery}, measured on validation responses, and \emph{storage recovery}, measured only after recovery by comparison with an open-weight checkpoint. We claim storage recovery only when entry-wise checkpoint evidence supports it.

\subsection{Contributions}\label{introduction-contributions}

\begin{enumerate}
\def\labelenumi{\arabic{enumi}.}
\item \textbf{C1 --- Separation (Section~\ref{the-attack-separation-by-observation}).} We apply known SiLU/GELU affine symmetry to the two-branch GLU product: Hessians supply gate-direction candidates, while outputs at \(x\) and \(-x\) separate magnitude/coupling and orientation subproblems.
\item \textbf{C2 --- Recovery (Section~\ref{the-attack-separation-by-observation}).} We construct the direction search, fit magnitudes and couplings, solve orientations, refine directions, and select storage scales.
\item \textbf{C3 --- Scaling (Section~\ref{nested-dimensional-decomposition-scaling-input-dimension-by-stages}).} Saved-residual repairs replace missing or duplicate candidates without new queries. Extensions based on Hessian-vector products replace quadratic full-Hessian storage with storage linear in input dimension for fixed probe, direction, and output counts. This makes the reported recovery over all 1,024 Qwen input coordinates and 128 output rows feasible.
\item \textbf{C4 --- Experimental results (Section~\ref{empirical-evaluation}).} Across all six high-precision Qwen layers, recovery over all 1,024 input coordinates and 128 output rows gives validation medians from \(3.9\times10^{-8}\) to \(8.8\times10^{-6}\). The high-precision Llama experiment recovers all 8,192 checkpoint-matched gate directions in the tested 256-coordinate subproblem, and the high-precision Gemma experiment covers all 1,152 input and output coordinates. In four experiments that execute the target in BF16, FP16, FP32, or FP32-stored TF32 arithmetic, median functional error ranges from \(0.0031\) to \(0.0463\), but none is storage-exact.
\end{enumerate}

\section{Background and related work}\label{background-and-related-work}
We review cryptanalytic weight recovery, GLU feed-forward blocks, and the precision measures used in our experiments.

\subsection{Cryptanalytic extraction of neural networks}\label{cryptanalytic-extraction-of-neural-networks}
Prediction-API model extraction often seeks a model that reproduces the target's behavior~\cite{Tramer16}. Model inversion instead uses model outputs to infer sensitive attributes or representative inputs~\cite{Fredrikson15}. Cryptanalytic extraction seeks the parameters themselves. Carlini, Jagielski, and Mironov~\cite{CJM20} recover fully connected ReLU networks at MNIST scale by locating inputs at which one hidden unit's preactivation is zero and estimating derivative differences across that unit's active/inactive boundary. Their construction uses the resulting slope discontinuity.

Later work retains this architecture but expands its activations and output interfaces. Their \emph{sign} and our \emph{orientation} denote the same binary \(\pm\) choice for a recovered direction, although the recovery procedures differ. Canales-Mart\'inez et al.~\cite{CanalesMartinez24} replace Carlini et al.'s exponential-time sign search with polynomial-time sign-recovery techniques and demonstrate extraction of a 1.2-million-parameter ReLU network. Foerster et al.~\cite{FMSH24} integrate Carlini et al.'s signature recovery with Canales-Mart\'inez et al.'s sign recovery, further optimize sign extraction, and identify signature recovery as the main runtime bottleneck. Qi et al.~\cite{QLWSW26} investigate PReLU, Leaky ReLU, HardTanh, ELU, and Step in both raw-output and hard-label settings, with procedures and results that depend on the activation and interface. Asselineau et al.~\cite{ADFM26} formalize affine-symmetric activations, including GELU and SiLU, and analyze the resulting sign ambiguity in chains of affine maps and componentwise activations. Their model does not include the parallel-branch product \(\sigma(W_gx)\odot(W_ux)\). We use the same activation symmetry in a different architecture: multiplication by the linear value branch turns input reversal into separate orientation-blind and orientation-linear GLU observations.

A separate line of work targets Transformer output layers rather than internal FFN or attention parameters. Carlini et al.~\cite{CPT24} reconstruct a Transformer language model's final embedding-projection matrix, up to a right-multiplication change of basis, from API outputs such as top-token log probabilities combined with logit bias. Its global change-of-basis ambiguity differs from the per-hidden-unit GLU scaling ambiguity resolved in Section~\ref{stage-2-and-stage-3-direction-refinement-and-storage-precision}.

\subsection{The Gated Linear Unit family}\label{the-gated-linear-unit-family}
The Gated Linear Unit was introduced by Dauphin et al.~\cite{Dauphin17} as a gating mechanism for convolutional language models. Shazeer~\cite{Shazeer20} later evaluated a family of GLU variants as Transformer FFN replacements. We test two variants:

\begin{itemize}
\item \textbf{SwiGLU} uses \(\sigma(z)=\mathrm{SiLU}(z)=z\cdot\mathrm{sigmoid}(z)\). The Qwen3 report specifies SwiGLU for the dense-model architecture used in Section~\ref{empirical-evaluation}~\cite{QwenTeam25}.
\item \textbf{GeGLU} uses a GELU gate. Shazeer's definition uses \(\sigma(z)=z\Phi(z)\)~\cite{Shazeer20}; the T5-v1.1-small checkpoint~\cite{T5v11Card} and Gemma target configuration in Section~\ref{empirical-evaluation} use their checkpoint and backend GELU implementations.
\end{itemize}

Both variants contain the two-branch product that is absent from the affine-map-plus-componentwise-activation class analyzed by Asselineau et al.; their work does not discuss GLUs~\cite{ADFM26}. Within the GLU form studied here, SwiGLU and GeGLU differ in the gate activation \(\sigma\). For our recovery method, this difference changes the activation-specific formulas but not the sequence of recovery stages: the Hessian decomposition uses \(\sigma'\) and \(\sigma''\), while the split between responses at \(x\) and \(-x\) uses identities specific to the chosen activation. Section~\ref{the-attack-separation-by-observation} develops these observations for SwiGLU, and Section~\ref{empirical-evaluation} evaluates GeGLU with the corresponding GELU formulas and probe scales.

\subsection{Storage precision}\label{storage-precision}
Section~\ref{empirical-evaluation} evaluates BF16, FP16, FP32, and FP32-stored TF32 targets. The structural stages recover a continuous, functionally equivalent block. Only Stage~3's scale selection and rounding depend on the declared storage format. Exact storage recovery requires every rounded entry to equal the corresponding stored target value.

We measure functional recovery by the validation error \(\|\hat f(x)-f(x)\|_2/\|f(x)\|_2\). For checkpoint evaluation, we normalize gate rows and greedily pair recovered and target rows one-to-one in descending absolute cosine. Gate cosine is the absolute cosine of a matched pair; orientation accuracy is the fraction with positive signed cosine; and magnitude error is the median relative error between paired gate-row norms. Hessian-vector-product (HVP) relative error is the Frobenius norm of the stacked prediction residual divided by the norm of the target stack. Joint-sign and power-of-two factor symmetries are aligned before value/output entry-wise scoring.

Stage~3 receives the target storage format but no checkpoint values. Reciprocal scaling of a recovered \(W_u\) row and matching \(W_o\) column preserves the continuous block function, so Stage~3 selects the scale with the smallest relative change on rounding. Equality and unit-in-the-last-place (ULP) distances are computed only after selection.

Unit-in-the-last-place (ULP) distance measures floating-point rounding gaps~\cite{Goldberg91}. At the same exponent, one FP32 ULP is \(2^{-16}\) of one BF16 ULP and \(2^{-13}\) of one FP16 ULP. A common ULP threshold is therefore not a comparable accuracy measure across these formats.

\input{figures/separation}

\section{The attack: separation by observation}\label{the-attack-separation-by-observation}
Figure~\ref{fig:pipeline} summarizes the fixed-dimension recovery. We first derive the Hessian and antipodal views, then use them to construct direction candidates and estimate magnitudes, couplings, and orientations.

The recovery runs in named stages. Stage~0 collects forward-query observations; Stage~1 produces gate-direction candidates; Stage~1A estimates magnitudes and couplings from the odd part; Stage~1B estimates orientations from the even part; Stage~2 refines directions and, in finite-precision experiments, polishes fixed-gate couplings; and Stage~3 selects the value/output scale and rounds to the declared storage format. Section~\ref{the-chain} introduces a conditional input-extension repair.

\subsection{Two structural views and their invariances}\label{two-structural-views-and-their-invariances}
The target is a bias-free GLU block \(f : \R^n \to \R^{n_{\mathrm{out}}}\) with input dimension \(n\), hidden width \(m\), and output dimension \(n_{\mathrm{out}}\),
\begingroup
\[
\begin{aligned}
y_k(x)
&= \sum_{\ell=1}^{m} W_o[k,\ell]\,
   \sigma\!\big(g_\ell(x)\big)\, u_\ell(x),\\
g_\ell(x) &:= w_{g,\ell}^\top x,
\qquad
u_\ell(x) := w_{u,\ell}^\top x ,
\end{aligned}
\]
\endgroup
with \(w_{g,\ell}, w_{u,\ell} \in \R^n\) the rows of \(W_g, W_u \in \R^{m \times n}\) and \(W_o \in \R^{n_{\mathrm{out}} \times m}\). Here \(\sigma\) is the gate activation. Using only queries to this isolated block, we seek its stored weights up to hidden-unit symmetries.

\textbf{View 1: the gate/cross-term Hessian decomposition.} Differentiating \(y_k\) twice in \(x\) and writing \(R_\ell := w_{g,\ell} w_{g,\ell}^\top\) and \(\Phi_\ell := w_{g,\ell} w_{u,\ell}^\top + w_{u,\ell} w_{g,\ell}^\top\), the per-output Hessian admits the exact decomposition
\[
\boxed{
\begin{aligned}
H_k(x) &\equiv \nabla_x^2 y_k(x)\\
&= \sum_{\ell=1}^{m} W_o[k,\ell] \Big(
  \sigma''(g_\ell(x))\, u_\ell(x)\, R_\ell
  + \sigma'(g_\ell(x))\, \Phi_\ell
\Big).
\end{aligned}}
\tag{1}
\]
Equation~(1) has no \(w_{u,\ell}w_{u,\ell}^\top\) term. The gate appears alone in the rank-1 term \(R_\ell\), while the value branch appears only in the rank-at-most-2 cross-term \(\Phi_\ell\). Each hidden unit's summed contribution is supported on \(\mathrm{span}\{w_{g,\ell},w_{u,\ell}\}\) and therefore also has matrix rank at most two. This is our structural comparison with the affine-map-plus-componentwise-activation class defined by Asselineau et al.~\cite{ADFM26}; it is not a claim that their attack uses our Hessian decomposition.

\textbf{View 2: the split under input sign reversal.} To separate the remaining quantities, substitute the odd/even SiLU decomposition from Section~\ref{glu-recovery-strategy} into the even and odd output parts \(E(x)\) and \(O(x)\) defined there. Using \(w_{g,\ell}^\top(-x) = -(w_{g,\ell}^\top x)\) and \((w_{u,\ell}^\top x)\,W_o[:,\ell] = C_\ell x\) gives
\[
\boxed{\begin{aligned}
E(x) &= \tfrac12 \sum_\ell (w_{g,\ell}^\top x)\,(C_\ell x),\\
O(x) &= \tfrac12 \sum_\ell h(w_{g,\ell}^\top x)\,(C_\ell x), \qquad h(z) = z\tanh\!\tfrac{z}{2}.
\end{aligned}}
\tag{2}
\]
The invariances in (2) are central to the attack:

\begin{itemize}
\item \textbf{\(O(x)\) is blind to the gate orientations.} \(h(z) = z\tanh(z/2)\) is \emph{even in its argument}, so \(h(w_{g,\ell}^\top x)\) is unchanged under \(w_{g,\ell}\mapsto -w_{g,\ell}\). The odd part depends only on gate magnitude/direction (through \(h\)) and on the coupling \(C_\ell\).
\item \textbf{\(E(x)\) is orientation-linear.} Substituting \(w_{g,\ell}=s_\ell c_\ell v_\ell\) gives
\[
E(x)=\tfrac12\sum_\ell s_\ell
 [\,c_\ell(v_\ell^\top x)(C_\ell x)\,].
\]
The bracket is orientation-free; only \(s_\ell\in\{\pm1\}\) carries orientation.
\end{itemize}

Table~\ref{tab:separation} summarizes which observation informs each recovery subproblem; it is the organizing principle of the attack.

\begin{table}[t]
\centering\scriptsize
\caption{Forward-query observations used for the per-hidden-unit recovery subproblems.}
\label{tab:separation}
\setlength{\tabcolsep}{4pt}%
\begin{tabular}{@{}p{0.14\columnwidth}p{0.11\columnwidth}p{0.27\columnwidth}p{0.33\columnwidth}@{}}
\toprule
Quantity & Obs. & Property & Estimator\\
\midrule
gate-direction candidate \(v_\ell\) & Hessian term \(R_\ell\) & \(R_\ell\) is positive semidefinite, rank~1 & rank-1 search (3.2), direction refinement (3.5)\\
coupling \(C_\ell\), magnitude \(c_\ell\) & \textbf{odd} part \(O(x)\) & \(O\) is orientation-blind & joint Levenberg--Marquardt (3.3)\\
orientation \(s_\ell\) & \textbf{even} part \(E(x)\) & \(E\) is linear in \(s\) & relaxed least squares (3.4)\\
\bottomrule
\end{tabular}
\end{table}

\textbf{Stage 0: constructing the observations.} Both views are estimated from forward queries. A \emph{probe} is a chosen base input \(x\), while a \emph{forward query} is one target evaluation at a specific input. A Gaussian probe has independent coordinates distributed as \(\mathcal{N}(0,\rho^2)\); \(\rho\) is the standard deviation of one input coordinate and controls how far \(W_gx\) moves from zero. We use \emph{(i) paired queries} at \(x\) and \(-x\), giving \(E,O\) with no derivatives, and \emph{(ii) four queries at symmetric perturbations of coordinate pair \((a,b)\)}:
\[
\resizebox{0.99\columnwidth}{!}{$\displaystyle
\widehat H_{ab}(x) = \frac{f(x{+}h e_a{+}h e_b) - f(x{+}h e_a{-}h e_b) - f(x{-}h e_a{+}h e_b) + f(x{-}h e_a{-}h e_b)}{4h^2}
$}
\]
Here \(h\) is the finite-difference step and \(e_a\) is coordinate vector \(a\). The same four calls give entry \((a,b)\) of every output Hessian \(\widehat H_k(x)\). Repeating over coordinate pairs forms the matrices used in Stage~1. The high-precision experiments use \(h \approx 3\times10^{-3}\); Section~\ref{empirical-evaluation} reports precision-specific changes. Section~\ref{observations-used-at-each-chain-step} uses directional differences when full coordinate resolution is unnecessary.

Different probe scales observe the near-zero, transition, or saturation regions of the gate response. Their numerical values depend on the activation, gate-row norms, and target and returned-output arithmetic. Table~\ref{tab:scales} gives the high-precision SwiGLU settings; Section~\ref{empirical-evaluation} reports other conditions. All observations come from forward responses.

\begin{table}[t]
\centering
\caption{Probe scales used for high-precision SwiGLU recovery; Section~\ref{empirical-evaluation} gives activation- and precision-specific changes.}
\label{tab:scales}
\small
\setlength{\tabcolsep}{4pt}%
\begin{tabular}{@{}>{\raggedright\arraybackslash}p{0.26\columnwidth} p{0.28\columnwidth} p{0.34\columnwidth}@{}}
\toprule
input-coordinate standard deviation \(\rho\) & observation / stage & purpose\\
\midrule
\(\approx 4\) & Hessian (Stage~1), even part (1B) & form direction candidates; estimate orientations\\
\(0.04\) & odd part, small inputs (1A) & keep gates near zero; estimate \(c_\ell^2C_\ell\)\\
\(30\) & odd part, large inputs (1A) & saturate \(\tanh\); separate \(c_\ell\) from \(C_\ell\)\\
\bottomrule
\end{tabular}
\end{table}

\subsection{Stage 1: gate-direction candidates from the Hessian subspace}\label{stage-1-gate-directions-via-rank-1-manifold-pursuit}
At a high level, Stage~0 converts query responses into estimated Hessian matrices. Stage~1 combines these matrices across probes and outputs to estimate a shared component subspace, then searches near that subspace for rank-1 matrices that provide gate-direction candidates.

More precisely, the gate appears alone in the rank-1 components \(R_\ell = w_{g,\ell} w_{g,\ell}^\top\) of (1). For a symmetric matrix \(A\), let \(\mathrm{vec_{sym}}(A)\in\R^{n(n+1)/2}\) list its upper-triangular entries, multiplying off-diagonal entries by \(\sqrt2\). Applying this representation to (1) gives
\[
\begin{aligned}
\mathrm{vec_{sym}}(H_k(x))
= \sum_\ell W_o[k,\ell]\big[&
 c^{(1)}_\ell(x)\,\mathrm{vec_{sym}}(R_\ell)\\[-2pt]
&+c^{(2)}_\ell(x)\,\mathrm{vec_{sym}}(\Phi_\ell)\big],
\end{aligned}
\tag{3}
\]
with \(c^{(1)}_\ell(x) = \sigma''(g_\ell(x)) u_\ell(x)\) and \(c^{(2)}_\ell(x) = \sigma'(g_\ell(x))\). Equation~(3) introduces no new target queries. It states that every exact vectorized Hessian is a linear combination of the same matrices \(R_\ell\) and \(\Phi_\ell\).

Let \(d\) be the number of Gaussian probes \(x_1,\ldots,x_d\). Each probe requires \(4n^2\) calls, one four-call difference per ordered coordinate pair; every call returns all \(n_{\mathrm{out}}\) outputs. With \(p=n(n+1)/2\), we form \(M_{\mathrm{obs}}\in\R^{(d n_{\mathrm{out}})\times p}\) by placing each \(\mathrm{vec_{sym}}(\widehat H_k(x_i))\) in one row. Section~\ref{algorithm-and-forward-query-cost} instantiates this count for the experimental dimensions.

The component vectors in (3) define the theoretical subspace
\[
\mathcal{V} := \mathrm{span}\{\mathrm{vec_{sym}}(R_\ell), \mathrm{vec_{sym}}(\Phi_\ell)\}_{\ell},
\]
where \(\mathrm{span}\) means the set of all linear combinations of the listed vectors. This subspace has dimension at most \(2m\). If the estimated Hessians equal the exact Hessians, every row of \(M_{\mathrm{obs}}\) lies in \(\mathcal V\). Finite-difference and rounding errors perturb the observed rows away from this subspace.

Stage~1 estimates a rank-\(2m\) subspace from \(M_{\mathrm{obs}}\), denoted by \(\widehat{\mathcal V}\). The inequality \(d\,n_{\mathrm{out}}\ge 2m\) is a necessary row-count heuristic for spanning \(2m\) components, not a sufficient identifiability condition: the response coefficients must also have adequate rank and conditioning.

\textbf{Searching for direction candidates.} The \(R_\ell\) components are positive-semidefinite (PSD) rank-1 matrices, whereas each cross-term \(\Phi_\ell\) is rank-2 indefinite when its gate and value rows are not collinear. This contrast motivates searching for PSD rank-1 matrices near \(\widehat{\mathcal V}\). It does not prove that arbitrary linear combinations in \(\mathcal V\) contain no other PSD rank-1 matrices; we therefore treat the search as an empirical recovery step and compare its candidates with checkpoints only after recovery. We score a candidate direction \(q\) by
\[
L(q; \widehat{\mathcal V})
:= \big\| \mathrm{vec_{sym}}(qq^\top) - P_{\widehat{\mathcal V}}\big(\mathrm{vec_{sym}}(qq^\top)\big)\big\|^2,
\quad q\in S^{n-1}.
\tag{4}
\]
Here \(P_{\widehat{\mathcal V}}\) returns the closest vector in \(\widehat{\mathcal V}\). A smaller value means that \(qq^\top\) better matches the shared Hessian structure; in exact arithmetic, \(q=\pm v_\ell\) gives zero. We minimize (4) from several initial directions with limited-memory Broyden--Fletcher--Goldfarb--Shanno (L-BFGS)~\cite{LiuNocedal89}. With \(r\) stored update pairs, L-BFGS uses \(O(nr)\) memory instead of an \(n\times n\) matrix. Its gradients are computed from \(\widehat{\mathcal V}\), not from the target. Multiple starts reduce failures at poor local minima.

After finding a direction, we remove its rank-1 component before the next search. Because deflation accumulates error, we rerun L-BFGS against the original \(\widehat{\mathcal V}\). Since \(q\) and \(-q\) give the same \(qq^\top\), two normalized candidates are duplicates if \(|q^\top q'|\) reaches 0.9995 in our experiments. We retain one from each pair and continue until \(m\) nonduplicate candidates remain. This known-width stopping rule establishes candidate count, not checkpoint-matched recovery; Section~\ref{empirical-evaluation} reports the latter where measured.

Stage~1 returns only candidate unit directions. Stages~1A and 1B estimate magnitudes and orientations.

\subsection{Stage 1A: magnitude and couplings from the odd part}\label{stage-1a-magnitude-and-couplings-from-the-odd-part}
Stage~1A estimates the per-hidden-unit magnitudes \(c_\ell\) and couplings \(C_\ell\) from the odd part \(O(x)\), with the directions \(v_\ell\) held fixed at their Stage-1 values. From (2),
\[
O(x) = \tfrac12 \sum_\ell h\big(c_\ell\,(v_\ell^\top x)\big)\,(C_\ell x),
\qquad h(z) = z\tanh\!\tfrac{z}{2},
\]
which is independent of the orientations \(s_\ell\). We fit \((c,W_u,W_o)\), with \(C_\ell=W_o[:,\ell]w_{u,\ell}^\top\), to paired-input observations at several probe scales.

Changing \(c_\ell\) changes the shape of \(\mathrm{silu}(c_\ell v_\ell^\top x)\), so it cannot be absorbed into a fixed rescaling of \(C_\ell\). Hessian residuals alone mix magnitude and coupling; observations at different probe scales separate them.

Write each probe as \(x=\rho\xi\), where \(\xi\sim\mathcal N(0,I)\). Then the gate argument is \(c_\ell\rho(v_\ell^\top\xi)\) and has standard deviation \(c_\ell\rho\); the linear factor is \(\rho C_\ell\xi\). Thus, \(\rho\) changes which part of the gate response the queries measure.

At a small probe scale chosen so that typical gate arguments satisfy \(|t|\ll1\), \(\tanh(t/2)=t/2+O(t^3)\), and hence \(h(t)=t^2/2+O(t^4)\). Substituting \(t=c_\ell\rho(v_\ell^\top\xi)\), the contribution of hidden unit \(\ell\) to \(O(\rho\xi)\) is approximately \(\tfrac14\rho^3(v_\ell^\top\xi)^2(B_\ell\xi)\), where \(B_\ell:=c_\ell^2C_\ell\). Only \(B_\ell\) appears, so rank-1 alternating least squares can estimate this product but cannot separate \(c_\ell\) from \(C_\ell\). At a larger probe scale chosen to move the gate arguments toward saturation, \(h(t)\approx|t|\); with \(B_\ell\) fixed, the same contribution is approximately \(\tfrac12\rho^2|v_\ell^\top\xi|(B_\ell/c_\ell)\xi\). The large-scale fit can therefore recover \(c_\ell\), then obtain \(C_\ell=B_\ell/c_\ell^2\). These estimates initialize a joint Levenberg--Marquardt update of \((c,W_u,W_o)\).

For \(N\) large-scale probes, a direct magnitude update would store an \(N\times n_{\mathrm{out}}\times m\) residual Jacobian. We instead compute Jacobian products and form a \emph{gate-coherence graph}: each vertex is a recovered hidden unit, with an edge between \(\ell\) and \(\ell'\) when \(|v_\ell^\top v_{\ell'}|\) exceeds a threshold. The solver updates one connected component at a time and recomputes the full residual between sweeps. For Qwen (\(m=3072\)) at threshold 0.45, most components contain one hidden unit and the largest contain about 100; this is an empirical Qwen property.

Stage~1A holds the Stage~1 directions fixed and estimates only magnitudes and couplings. Stage~1B next estimates orientations, after which Stage~2 refines the directions against the Hessian observations.

\subsection{Stage 1B: orientations from the even part}\label{stage-1b-signs-from-the-even-part}
Using the magnitudes and couplings from Stage~1A and the directions from Stage~1, Stage~1B estimates the gate orientations \(s_\ell\) from the even part \(E(x)\). From (2),
\[
E(x) = \tfrac12 \sum_\ell s_\ell\,\big[\,c_\ell (v_\ell^\top x)(C_\ell x)\,\big],
\]
which is \emph{linear} in the orientation vector \(s\in\{\pm1\}^m\). Stacking the (known) brackets over probes and outputs into a design matrix \(\Psi\) and the observed even parts into \(e\), the orientations solve a single global least-squares
\[
\min_{s\in\{\pm1\}^m} \|\Psi s - e\|^2,
\]
relaxed to \(s\in\R^m\) and rounded. Solving the orientation-blind couplings first avoids the circular error of alternating orientation and coupling updates.

\subsection{Stage 2 and Stage 3: direction refinement and storage precision}\label{stage-2-and-stage-3-direction-refinement-and-storage-precision}
Stage~2 fixes magnitudes, orientations, and couplings while L-BFGS minimizes the Hessian residual over \(v_\ell\). It returns to the observations from which Stage~1 formed the direction candidates rather than asking the Stage~1A objective to adjust them.

In the four finite-precision runs, Stage~2 is followed by a bounded, fixed-gate factor polish. For fixed \(w_{g,\ell}\), let \(\mathcal L_\ell(C_\ell)\) map the rank-1 coupling \(C_\ell=W_o[:,\ell]w_{u,\ell}^{\top}\) to that unit's stacked Stage~0 Hessian prediction. Saved-residual projections onto each unit's gate/value Hessian basis rank the units. For a selected unit, we subtract the others' predictions and minimize the complete Hessian residual over rank-1 \(C_\ell\). Whitening by \(\mathcal L_\ell^*\mathcal L_\ell\) reduces this local problem to a best rank-1 approximation, given by its leading singular triplet. We retain the update only if the complete residual falls. This polish uses saved observations, makes no target queries, and cannot change gate directions.

Stage~3 selects a functionally equivalent value/output scale and rounds it to the declared storage format. Let \(Q_{\mathrm{st}}\) denote entry-wise rounding to that format, and let \(a_\ell,b_\ell\) denote the recovered \(W_u\) row and matching \(W_o\) column. Because \((a_\ell,b_\ell)\mapsto(\alpha_\ell a_\ell,b_\ell/\alpha_\ell)\) preserves the continuous block function, we center the scale search at \(\alpha_{\ell,0}=\sqrt{\lVert b_\ell\rVert_2/\lVert a_\ell\rVert_2}\). In Section~\ref{native-execution-paths}, we test 513 logarithmically spaced values in \([\alpha_{\ell,0}/\sqrt{2},\sqrt{2}\alpha_{\ell,0}]\), scored by
\[
\resizebox{0.88\columnwidth}{!}{$r_\ell(\alpha)=\frac{\lVert \alpha a_\ell-Q_{\mathrm{st}}(\alpha a_\ell)\rVert_2^2}{\lVert \alpha a_\ell\rVert_2^2}+\frac{\lVert b_\ell/\alpha-Q_{\mathrm{st}}(b_\ell/\alpha)\rVert_2^2}{\lVert b_\ell/\alpha\rVert_2^2}.$}
\]
Stage~3 chooses the minimum-score candidate, rounds \(a_\ell,b_\ell\) at that scale, and rounds \(W_g\) directly. The score uses no checkpoint values.

\section{Scaling and conditional repair}\label{nested-dimensional-decomposition-scaling-input-dimension-by-stages}
The fixed-dimension recovery may begin on selected input coordinates and output rows. This section repairs incomplete direction sets and expands that coverage.

For more output rows, Stage~0 retains more coordinates from each returned vector before Stages~1--3 run on the wider slice. For more input coordinates, we run Stages~0--2 on the starting subspace, then hold the recovered columns fixed while adding columns of \(W_g,W_u\). After reaching the full input dimension, we apply forward polish and validate before Stage~3.

Explicit Hessian storage grows as \(O(d n_{\mathrm{out}}n^2)\). With \(d=64\) and FP64 upper triangles, a 128-row slice requires about 32~GiB at Qwen's \(n=1024\) and 128~GiB at Llama's \(n=2048\); all 1,024 Qwen rows require about 256~GiB. This motivates storage that is linear in \(n\).

We use a nested input-dimension chain: recover a small input subspace, then keep its columns fixed while adding new ones. Extensions estimate Hessian-vector products instead of full Hessians, so storage grows linearly with \(n\) when probe, direction, and output counts are fixed.

Alongside this dimensional expansion, saved-observation residuals and other attacker-available diagnostics determine whether a conditional repair runs. Table~\ref{tab:playbook} distinguishes these repairs and the input-extension steps.

Stage~1 uses two distinct direction-set repairs for incomplete results. \emph{Residual-subspace rescue} is triggered when the candidate count is below the known hidden width \(m\); it initializes rank-1 pursuit from unexplained energy in the saved Stage~0 Hessian basis. \emph{Leave-one-out deflation} instead handles a near-duplicate pair: it deflates every candidate except the suspected duplicate and reruns rank-1 pursuit to seek a replacement. Both repairs use saved Stage~0 observations and require neither checkpoint weights nor new target queries.

\begin{table*}[t]
\centering
\caption{Recovery and extension playbook. Each action uses query-based diagnostics; checkpoint weights are used only to evaluate the final recovery.}
\label{tab:playbook}
\small
\setlength{\tabcolsep}{4pt}%
\begin{tabular}{@{}p{0.14\textwidth}p{0.24\textwidth}p{0.24\textwidth}p{0.26\textwidth}@{}}
\toprule
point in recovery & obstacle & attacker-available diagnostic & playbook action\\
\midrule
Stage~1 candidates & too few candidates & candidate count below known \(m\); residual Hessian energy & residual-subspace rescue\\
Stage~1 candidates & near-duplicate pair & pairwise absolute cosine & leave-one-out deflation\\
conditional extension repair (Stage~1X) & warm extension loses a reliable prefix & residual with prefix fixed; mixed prefix/extension Hessian fit & solve new \(W_g,W_u\) columns with the prefix fixed\\
Stage~1A/1B & magnitude/coupling/orientation ambiguity & odd/even residuals; orientation-solve conditioning & joint odd-part solve; linear even-part orientation solve\\
Stage~2 & direction-limited functional floor & validation forward/Hessian residual & refine directions on saved Hessians\\
After Stage~2 & concentrated value/output residual & per-unit saved-residual projection score & update selected rank-1 couplings with gates fixed\\
Between chain steps & large variation in value/output factor scales & spread of per-unit \(\|w_{u,\ell}\|/\|W_o[:,\ell]\|\) & query-free per-unit scale normalization\\
Input-chain step & unstable large chunks; quadratic Hessian storage & extension-step residual & Hessian-vector products; bounded-\(n_{\mathrm{ext}}\) extension\\
After full \(n\) & remaining full-input error & fresh forward/HVP validation & apply forward polish; retain the selected result; apply Stage~3\\
\bottomrule
\end{tabular}
\end{table*}

\subsection{The nested input-dimension chain}\label{the-chain}
Choose \(0 < n_1 < \ldots < n_T = n\) with \(n_{i+1} - n_i \le n_{\mathrm{ext}}^{\max}\), a per-model bound reported in Section~\ref{empirical-chain-settings-and-open-questions}.

\textbf{Base step.} Apply Stages~0--2 to probes in \(\R^{n_1}\), zero-padded to \(\R^n\), to form and refine \(m\) candidate units on the starting subspace.

\textbf{Extension step.} Hold prior columns fixed, recover the next \(n_{\mathrm{ext}}\) columns of \(W_g,W_u\), and update \(W_o\) using probes over \(\R^{n_{i+1}}\). If this warm extension is ill-conditioned despite a reliable prefix, Stage~1X queries mixed prefix/extension Hessian blocks at probes whose extension coordinates are zero. Across output slices, these blocks form a ridge-regularized linear system for the new columns with the prefix fixed.

Each extension introduces \(2mn_{\mathrm{ext}}\) new entries across the gate and value branches. Section~\ref{empirical-chain-settings-and-open-questions} reports the model-specific schedules.

\textbf{Step size.} The starting width \(n_1\) and each extension width are empirical, model-specific choices reported in Section~\ref{empirical-chain-settings-and-open-questions}.

\textbf{Termination.} After the chain reaches \(n_T=n\), apply forward polish and validate the full-input recovery. Stage~3 then selects the scale splits and rounds the result to the declared storage format.

\subsection{Hessian observations at each chain step}\label{observations-used-at-each-chain-step}
The initial \(n_1\)-coordinate step uses the coordinate-wise finite differences from Section~\ref{two-structural-views-and-their-invariances} to estimate every Hessian entry within that input subspace. Each extension instead estimates Hessian-vector products \(H_k(x)v\). For coordinate \(a\), the corresponding entry is
\[
\begin{aligned}
\big[H_k(x)v\big]_a \approx \frac{1}{4h^2}\big(&f_k(x+h e_a+h v)-f_k(x+h e_a-h v)\\
&-f_k(x-h e_a+h v)+f_k(x-h e_a-h v)\big).
\end{aligned}
\]
Each call returns all \(n_{\mathrm{out}}\) outputs. For one \((x,v)\) pair, repeating over \(a=1,\ldots,n\) uses \(4n\) calls and forms an \(n_{\mathrm{out}}\times n\) array. A full Hessian would use \(4n^2\) calls and form an \(n_{\mathrm{out}}\times n\times n\) array. Each additional \((x,v)\) pair costs another \(4n\) calls; Section~\ref{algorithm-and-forward-query-cost} gives the totals.

\subsection{Full-input refinement}\label{the-chain-at-deployment-input-dimension}
\textbf{Scale normalization.} The function-preserving branch scale can poorly condition extension optimization. Before the next extension, we apply
\[
\resizebox{0.95\columnwidth}{!}{$\displaystyle
\alpha_\ell=\sqrt{\|W_o[:,\ell]\|/\|w_{u,\ell}\|},\qquad (w_{u,\ell},W_o[:,\ell])\mapsto(\alpha_\ell w_{u,\ell},W_o[:,\ell]/\alpha_\ell),
$}
\]
which equalizes the two factor norms and requires no target queries.

\textbf{Forward polish.} At width \(n_i\), Gaussian probes cover its first \(n_i\) coordinates and zero-pad the rest. Two thirds of the saved query pairs define the forward-error objective (\emph{training} pairs); the remaining third are held out to detect overfitting (\emph{guard} pairs). L-BFGS updates units ranked by the gradient norm of this saved-response objective, while separate fresh forward and HVP responses select among candidates. The Qwen schedule starts with 64 units and 14k/7k training/guard probes, then grows to all 3,072 units and 80k/40k probes; Table~\ref{tab:fulln} reports the call counts. It stops at median \(\le10^{-6}\) and p99 \(\le2\times10^{-6}\), or at the schedule limit.

\subsection{Empirical chain settings and open questions}\label{empirical-chain-settings-and-open-questions}
We chose chain widths empirically: Qwen uses at most 192 new coordinates per step, Llama starts at 256 and adds 128, and Gemma starts at 256 and adds 64. Qwen layer~5 required a separate repaired path after its original chain plateaued. These schedules were not a common sweep and do not establish a general step-size rule.

\section{Algorithm and forward-query cost}\label{algorithm-and-forward-query-cost}
The counts below measure isolated-block queries, not model-API calls. A token API does not expose chosen block inputs or block outputs, and we do not show how to derive one block response from final model outputs. Reusing a saved response adds local computation but no call.

\textbf{Base budget.} At \(n_1=192,d=64\), coordinate Hessians require \(4dn_1^2=9{,}437{,}184\) calls. Stage~1A/1B uses 440,000 paired calls and candidate selection uses 4,096 responses, producing a 9,881,280-call attack transcript. Table~\ref{tab:native-precision} adds a shared 4,096-query factor-polish selector to each row and excludes a separate 4,096-query final evaluation. Precision-specific searches are also included; Kshana's triggered branch adds 28,096 calls. Saved-residual repairs, Stage~2, and factor polish make no target calls.

\textbf{Full-input budget.} Qwen extends 192 coordinates through endpoints 384, 576, 768, 960, and 1024. With \(M\) base probes and \(K\) directions per probe, a step ending at \(n_i\) costs \(4MKn_i\) calls. At \(M=8,K=64\), extension costs 7,602,176 calls and raises the pre-polish total to 17,483,456. An explicit \(n=1024,d=64\) coordinate Hessian would cost 268,435,456 calls.

\textbf{Forward-polish budget.} Table~\ref{tab:fulln} reports 477,000--986,000 training/guard calls. Each candidate comparison uses 1,048,576 HVP calls plus 4,096 forward responses shared across the candidates in that comparison. Ten comparisons therefore add 10,526,720 calls and make the full path 28,487,176 calls for Qwen layers~1, 3, and 4; the table excludes these comparison calls.

\textbf{Compute and memory.} Base storage scales as \((d,n_{\mathrm{out}},n_1,n_1)\), while extension storage is linear in the current input dimension. Output width therefore trades compute and memory at a fixed call count.

\section{Empirical evaluation}\label{empirical-evaluation}
We first develop the recovery playbook with FP64 target computation and returned outputs, which preserve response differences that finite-precision rounding can erase. We then test FP64 transfer to Llama and Gemma, four finite-precision target configurations, output-only rounding, and Qwen BF16 execution.

\subsection{High-precision Qwen recovery experiments}\label{forward-query-attack-on-qwen3-0.6b-swiglu-ffn-blocks}
These experiments use \path{Qwen/Qwen3-0.6B}~\cite{Qwen3Card}. The cited artifact points to revision \texttt{c1899de289a0}; weights are cast to FP64.

\textbf{Recovered dimensions.} Table~\ref{tab:perlayer} evaluates recovery of the \(m=3072\) hidden units on 192 input coordinates and 128 output rows without checkpoint initialization. Table~\ref{tab:fulln} extends these estimates to \(n=1024\) over the base 128 output rows and then applies forward polish.

\textbf{Output slice.} Stage~1 uses \(d n_{\mathrm{out}}\ge2m\) as a row-count heuristic. Our setting \((d,n_{\mathrm{out}})=(64,128)\) satisfies it, since \(d n_{\mathrm{out}}=8192\) and \(2m=6144\). Using 256 output rows also returns 3,072 nonduplicate candidates but doubles the Stage~0 observation storage. We score only the 128 recovered rows of \(W_o\).

\begin{table}[H]
\centering\footnotesize
\caption{High-precision Qwen recovery at \((n_1,m,n_{\mathrm{out}})=(192,3072,128)\), before forward polish. Stage~1A small probes use \(\rho=0.04\); Table~\ref{tab:scales} lists the run's other probe scales.}
\label{tab:perlayer}
\setlength{\tabcolsep}{3pt}%
\begin{tabular}{@{}lrr@{}}
\toprule
layer & mag err $\downarrow$ & fwd med $\downarrow$\\
\midrule
L0 & 0.17\% & \(8.2\times10^{-4}\)\\
L1 & 0.18\% & \(1.1\times10^{-3}\)\\
L2 & 0.20\% & \(1.4\times10^{-3}\)\\
L3 & 0.13\% & \(8.7\times10^{-4}\)\\
L4 & 0.16\% & \(1.0\times10^{-3}\)\\
L5 & 0.17\% & \(8.9\times10^{-4}\)\\
\bottomrule
\end{tabular}
\par\medskip
\caption{Full-input high-precision Qwen recovery after chaining and forward polish on 128 output rows. Forward and HVP errors use fresh target responses without checkpoint weights. L5 is set apart because its chain required the separate repaired path noted in Section~\ref{empirical-chain-settings-and-open-questions}.}
\label{tab:fulln}
\setlength{\tabcolsep}{3pt}%
\begin{tabular}{@{}lrrrr@{}}
\toprule
layer & fwd med $\downarrow$ & fwd p99 $\downarrow$ & HVP rel $\downarrow$ & training/guard calls\\
\midrule
L0 & \(7.53\times10^{-8}\) & \(1.16\times10^{-7}\) & \(3.48\times10^{-7}\) & 478.5k\\
L1 & \(7.11\times10^{-7}\) & \(1.25\times10^{-6}\) & \(2.69\times10^{-6}\) & 477.0k\\
L2 & \(3.91\times10^{-8}\) & \(8.49\times10^{-8}\) & \(2.11\times10^{-7}\) & 986.0k\\
L3 & \(1.15\times10^{-7}\) & \(1.91\times10^{-7}\) & \(7.83\times10^{-7}\) & 477.0k\\
L4 & \(1.21\times10^{-7}\) & \(2.37\times10^{-7}\) & \(3.95\times10^{-7}\) & 477.0k\\
\midrule
L5 & \(8.84\times10^{-6}\) & \(1.72\times10^{-5}\) & \(5.60\times10^{-5}\) & 657.0k\\
\bottomrule
\end{tabular}
\end{table}

Across six layers, comparison with the checkpoint gives 100\% orientation accuracy and 0.13--0.20\% magnitude error. Validation error is \(8.2\times10^{-4}\)--\(1.4\times10^{-3}\).

\textbf{Repair triggers and probe-scale sensitivity.} Stage~1 returned fewer candidates than the known width \(m\) on L2/L4/L5, invoking residual-subspace rescue; a duplicate on L5 also invoked leave-one-out deflation. Both repairs use saved observations and make no new target queries. At fixed query budget, the Qwen base experiment tests \(\rho\in\{0.12,0.08,0.04,0.02\}\). The respective magnitude errors are 0.39\%, 0.256\%, 0.194\%, and 0.186\%, with little gain below \(\rho=0.04\).

\textbf{Full input dimension.} The chain extends each layer to \(n=1024\) over the base 128 output rows. Forward polish produces several candidates. Fresh query-only evaluation uses random inputs over all 1024 coordinates and finite-difference HVP probes to select among them without checkpoint weights. Table~\ref{tab:fulln} reports this evaluation for the selected candidates.

For these full-input recoveries, comparison with the checkpoint places 84--97\% of each of \(W_g\), \(W_u\), and the 128 recovered rows of \(W_o\) within one ULP of Qwen3-0.6B's BF16 storage grid. This rate covers all entries, including near-zero ones, and is therefore not comparable with the significant-entry rates reported for Gemma in Section~\ref{real-geglu-gemma}.

\subsection{A second model family: Llama-3.2-1B}\label{second-model-family-llama}
We test layer~2 of \path{unsloth/Llama-3.2-1B}~\cite{Llama32Card}. The cited artifact points to revision \texttt{9535bd9b1d1d}; its geometry is \((n_1,m,n_{\mathrm{out}})=(256,8192,128)\). BF16-stored weights are computed and returned in FP64. Checkpoint evaluation matches all 8,192 candidates with minimum cosine 0.99993 and median 1.000. Orientation accuracy is 100\%, magnitude error is 0.62\%, and median validation error is \(2.0\times10^{-3}\). This is functional transfer at \(2.7\times\) Qwen's hidden width, not storage exactness.

For Stage~1X, a warm extension from \(n=256\) to 512 loses prefix alignment. The conditional mixed-Hessian solve instead reaches \(9.7\times10^{-4}\) validation error. After checkpoint-free BF16 rounding, 81.1\% of all 9,437,184 recovered entries---\(2\times8192\times512\) gate and value entries plus \(128\times8192\) output entries---are exact, 95.8\% are within one ULP, and 97.6\% are within two ULP. This is a partial-input extension, not a full Llama chain.

\subsection{A second activation: GeGLU on Gemma-3-1B}\label{real-geglu-gemma}
We test activation transfer on Gemma-3-1B layer~2. The target checkpoint is \path{unsloth/gemma-3-1b-pt}~\cite{Gemma3Card}. Its GeGLU block uses the checkpoint's tanh-approximated GELU:
\[
\begin{aligned}
f_{\mathrm{Ge}}(x)&=W_o\!\left(\sigma_{\mathrm{Ge}}(W_gx)\odot W_ux\right),\\
\sigma_{\mathrm{Ge}}(z)&=\tfrac{z}{2}\!\left[1+\tanh a(z)\right],
\quad
a(z)=\sqrt{\tfrac{2}{\pi}}\left(z+0.044715z^3\right).
\end{aligned}
\]
SwiGLU instead uses \(\sigma_{\mathrm{Swi}}(z)=\tfrac{z}{2}[1+\tanh(z/2)]\). Both activations satisfy \(\sigma(z)-\sigma(-z)=z\), so the even observation \(E(x)\) and Stage~1B orientation solve are unchanged. The orientation-blind odd kernels used by Stage~1A differ:
\[
h_{\mathrm{Swi}}(z)=z\tanh(z/2),
\qquad
h_{\mathrm{Ge}}(z)=z\tanh a(z).
\]
Near zero, \(h_{\mathrm{Ge}}(z)=\sqrt{2/\pi}\,z^2+O(z^4)\), rather than \(h_{\mathrm{Swi}}(z)=z^2/2+O(z^4)\), and the two kernels enter saturation at different gate arguments. The Hessian model likewise replaces the SiLU derivatives by \(\sigma_{\mathrm{Ge}}'\) and \(\sigma_{\mathrm{Ge}}''\).

The reported Gemma configuration retains the Stage~0 finite-difference step \(h=0.003\) and uses \(\rho=4.071\) for its base Hessian probes, but changes the Stage~1A small/large probe scales from the high-precision SwiGLU values \(0.04/30\) to \(0.12/15\). The full-input chain uses \(\rho=0.8\) for its Hessian-vector probes instead of Qwen's \(2.0\). These empirical scales place the small probes in the quadratic regime used to estimate \(B_\ell=c_\ell^2C_\ell\), the large probes near saturation to separate \(c_\ell\) from \(C_\ell\), and the chain probes below the over-saturated regime.

The geometry is \((n,m,n_{\mathrm{out}})=(1152,6912,1152)\), with FP64 computation and outputs over BF16-stored weights. The pipeline returns 6,912 candidates and covers all input/output coordinates. Full-input HVP error is \(8.4\times10^{-3}\); the forward median, p99, and maximum are \(4.9\times10^{-3}\), \(1.8\times10^{-2}\), and \(5.6\times10^{-2}\). On entries at least 5\% of their row maximum, exact/within-one/within-two-ULP rates are 63.3/91.3/96.4\% for \(W_g\), 61.5/89.1/95.4\% for \(W_u\), and 59.0/90.7/96.7\% for \(W_o\); the recovery is not storage-exact.

\subsection{Target computation precision}\label{native-execution-paths}
Table~\ref{tab:native-precision} reports four target configurations. Because model width and activation also differ, the rows do not isolate precision.

Qwen and Kshana use BF16 and FP16 inputs, intermediate projections, fused activation/product values, and returned outputs in the vLLM inference backend. T5 uses an FP32 reference path. AMD-Llama stores FP32 weights and uses TF32 matrix multiplications with FP32 elementwise and returned values in vLLM. The reported Stage~2 and factor-polish computations use attacker-side FP32 with TF32 disabled; each local rank-1 solve uses FP64 eigendecomposition and singular-value decomposition.

\begin{table*}[!t]
\centering\footnotesize
\caption{Reported target configurations after factor-polish selection. Geometry is \((n,m,n_{\mathrm{out}})\). Attack calls include the shared 4,096-query selector but exclude the disjoint 4,096-query final evaluation reported here. Gate cosine is the absolute cosine after matching recovered and checkpoint gate rows; 1 denotes the same direction.}
\label{tab:native-precision}
\setlength{\tabcolsep}{2.2pt}%
\begin{tabular}{@{}llrrrr@{}}
\toprule
target & storage / target arithmetic & geometry & attack calls & \shortstack{validation\\med./p90 \(\downarrow\)} & \shortstack{gate cosine\\med./min \(\uparrow\)}\\
\midrule
Qwen3-0.6B L0 (SwiGLU) & BF16 / BF16 vLLM & 192,3072,512 & 9.885M & .046269/.077392 & .999946/.051573\\
Kshana-170M~\cite{Kshana170M} L0 (SwiGLU) & FP16 / FP16 vLLM & 192,1536,576 & 9.913M & .008738/.017676 & .999987/.038181\\
T5-v1.1-small L0 (GeGLU) & FP32 / FP32 reference & 192,1024,512 & 10.625M & .003056/.004604 & .999999/.997877\\
AMD-Llama-135M~\cite{AMDLlama135M} L0 (SwiGLU) & FP32 / TF32 vLLM & 192,2048,768 & 10.623M & .008651/.019833 & .999997/.260373\\
\bottomrule
\end{tabular}
\end{table*}

\textbf{Final selection.} Stage~1 keeps \(m\) candidates satisfying its distance and duplicate thresholds. Stage~2 selects the current or alternative \(m\)-candidate branch by saved-observation residual. After the fixed-gate factor polish and Stage~3 rounding, a shared fresh 4,096-query set selects the smallest median-error candidate whose p90 error is no more than 1\% above the incumbent's; the incumbent remains unless the median improves by at least 0.1\%. The selected configurations polish 24 units for Qwen, Kshana, and T5, and eight for AMD-Llama. The disjoint 4,096-query evaluation in Table~\ref{tab:native-precision} runs only after this choice.

The \path{google/t5-v1_1-small} run~\cite{T5v11Card} tests FP32 GeGLU. Query-only searches choose \(\rho=0.3,h=0.0375\). Removing one near-duplicate leaves 1,024 candidates on 192 input coordinates. Checkpoint direction cosine has median \(0.9999994\) and minimum \(0.997877\), with correct orientation for every match. Stage~2 lowers the saved residual from 0.03795 to 0.005192, and fixed-gate factor polish lowers it further to 0.005166. The disjoint evaluation over all 512 outputs gives median/p90 error 0.003056/0.004604.

On the disjoint evaluation, the selected polish lowers both the median and p90 error in all four configurations relative to their unpolished incumbents. The relative median reductions range from 0.4\% to 3.8\%, and the relative p90 reductions range from 0.2\% to 9.6\%. These four single-run results show a modest functional improvement from the bounded polish; they do not establish a general repair guarantee. None of the four final recoveries reproduces every stored weight. ULP evaluation supports that conclusion but is not used to compare formats.

Worst-case direction quality also differs. Median matched gate cosine exceeds \(0.9999\) in every row, but the minimum falls to 0.051573 for Qwen BF16, 0.038181 for Kshana FP16, and 0.260373 for AMD-Llama TF32; only the FP32 T5 run keeps every matched cosine above 0.99. A median near 1 therefore shows that most matched pairs align closely, not that every checkpoint gate direction is recovered. Because the factor polish fixes \(W_g\), it cannot change these direction statistics.

\subsection{Rounding only the returned output}\label{rounded-output-recovery}
To isolate returned-output rounding, this experiment computes Qwen3-0.6B layer~0 in FP64 and rounds only the returned vector at \((n_1,m,n_{\mathrm{out}})=(192,3072,128)\); we call its format the \emph{returned channel}. FP32 uses \(h=0.3\), BF16 uses \(h=5\), and FP16 combines \(h=5,10\) estimates by Richardson extrapolation~\cite{Richardson11}. The relaxed BF16 diagnostic raises the Stage~1 direction-acceptance threshold from 0.02 to 0.20. Saved residual and candidate count select the construction; later stages are unchanged.

\begin{table}[t]
\centering\small
\caption{Recovery with rounded returned vectors. The candidate column reports retained nonduplicate candidates relative to known width, not checkpoint-matched direction coverage. Errors use the same FP64 block and 4,096 probes; BF16 is a relaxed diagnostic.}
\label{tab:precision-recovery}
\setlength{\tabcolsep}{3.5pt}
\begin{tabular}{@{}lrrrr@{}}
\toprule
channel & candidates & residual & median & max\\
\midrule
FP32 & $3071/3072$ & 0.02585 & \textbf{0.00805} & 0.16441\\
FP16 Richardson & $3072/3072$ & 0.03570 & 0.01949 & 0.19772\\
BF16 relaxed & $2975/3072$ & 0.13929 & 0.08039 & 0.22890\\
\bottomrule
\end{tabular}
\end{table}

FP32 reaches $0.805\%$ median functional error with 3,071 retained candidates. FP16 retains the known width of 3,072 candidates and reaches $1.949\%$. Relaxed BF16 improves to $8.039\%$ with 2,975 candidates after three bounded residual-subspace rescue rounds; strict BF16 recovery is therefore not supported.

Adding the multi-scale Stage~1A fit gives the largest improvement across channels (Appendix Table~\ref{tab:precision-modules}). Residual-subspace rescue increases the candidate count but does not always lower the saved residual.

\subsection{Qwen BF16 execution stress test}\label{native-qwen-bf16}
BF16/vLLM execution rounds intermediate projections as well as the returned output. We ran Qwen3-0.6B layer~0 at the same base input dimension and probe count, retaining the first 512 output rows. The attack uses the common 9,881,280 calls plus 4,096 factor-polish selection calls; the disjoint 4,096-query final evaluation is excluded. We did not increase the observation budget to seek exact BF16 weights.

On these output rows, the run returns 3,072 nonduplicate direction candidates. Factor polish lowers the saved-observation residual from 0.156373 to 0.156341; the disjoint validation median/90th-percentile errors are 0.046269/0.077392. Checkpoint-based evaluation gives median/minimum matched gate cosine 0.999946/0.051573, so candidate-set completeness does not establish recovery of every target direction. Across all 2,752,512 recovered weights, 2.19\% match the checkpoint exactly, 6.38\% are within one ULP, and 10.33\% are within two ULP.

\section{Limitations and defensive implications}\label{limitations-and-defensive-implications}
The experiments require direct, full-vector queries to a bias-free isolated GLU block and correct architecture metadata. Qwen covers 1,024 input coordinates but 128 of its 1,024 output rows; Llama covers a 256-coordinate base and a diagnostic extension to 512 of its 2,048 input coordinates; Gemma covers all 1,152 input/output coordinates. Each target has one run, the finite-precision configurations are not controlled comparisons, and validation uses Gaussian block inputs. A token API does not expose the required interface. The results therefore do not establish model-API feasibility, success probability, precision causality, or downstream task equivalence.

The observed query volumes and sensitivity to returned precision suggest conditional mitigations for any service that exposes a comparable block interface: restrict such access, rate-limit or detect structured antipodal and finite-difference probes, and reduce returned precision where the application permits. These defenses are not evaluated here, and output quantization did not eliminate functional recovery in our tests. At 10--100\,ms per serialized call, 9.9--28.5 million block queries would take roughly 27 hours to 33 days before quotas; this illustration is not a measured model-API cost.

\section{Conclusion}\label{conclusion}
We presented a forward-query method for isolated bias-free GLU feed-forward blocks. Finite-difference curvature supplies candidate gate directions; outputs at \(x\) and \(-x\) separate magnitude and coupling estimation from orientation estimation. The experiments demonstrate functional recovery across the reported SwiGLU and GeGLU settings, but none of the finite-precision runs is storage-exact.

Applying this method to a complete language model still requires a query-only procedure that handles attention, residual connections, and preceding blocks using only final model outputs. Developing such a procedure remains the principal open step toward an end-to-end attack.

\subsection*{Ethics}
We evaluated public checkpoints locally and queried no proprietary service. The method is dual use: it can inform defenses for unusually rich inference interfaces, but could also facilitate extraction from a service that exposes an isolated-block equivalent. Because no such end-to-end attack is demonstrated, we state that boundary explicitly and defer code release until publication.

\appendix
\section{Recovery playbook pseudocode}
\begin{algorithm}[H]
\caption{Isolated-GLU recovery overview}
\label{alg:playbook}
\begin{algorithmic}[1]\footnotesize
\Require Isolated bias-free GLU function \(f\); returned-output operator \(Q_{\mathrm{out}}\); storage-rounding operator \(Q_{\mathrm{st}}\); query budgets
\State collect finite-difference and antipodal observations through \(Q_{\mathrm{out}}(f(x))\)
\State search Hessian observations for \(v_\ell\); repair candidates from saved residuals
\State calibrate \(c_\ell,C_\ell\) from odd observations; solve \(s_\ell\) from even observations
\State refine directions; optionally polish fixed-gate rank-1 couplings on saved observations
\If{a full input dimension is required}
\State extend columns, normalize branch scales, and polish on saved responses
\State select the candidate by fresh forward/HVP validation
\EndIf
\State select scale splits and apply \(Q_{\mathrm{st}}\); return the recovered weights
\end{algorithmic}
\end{algorithm}

\section{Recovery-module comparison under rounded outputs}
\begin{table}[H]
\centering\small
\caption{Stage~2 residual by recovery configuration and returned-output format (lower is better). The residual-repair row reruns Stage~2 after rescue.}
\label{tab:precision-modules}
\setlength{\tabcolsep}{4pt}
\begin{tabular}{@{}lrrr@{}}
\toprule
configuration & FP32 & FP16 & BF16\\
\midrule
Stage~1 fixed directions & 0.59853 & 0.67801 & 0.63383\\
Stage~2 direction refinement & 0.45939 & 0.61116 & 0.50732\\
\shortstack[l]{Residual repair\\+ Stage~2} & 0.46669 & 0.60572 & 0.47740\\
Multi-scale Stage~1A + Stage~2 & \textbf{0.02585} & \textbf{0.03570} & \textbf{0.13929}\\
\bottomrule
\end{tabular}
\end{table}

\section{Base-stage reproduction recipe}
For Qwen3-0.6B layer~0, run these scripts in order:
\begin{enumerate}
\item \path{qwen_forward_blackbox_observe.py};
\item \path{qwen_forward_blackbox_base_recover.py};
\item \path{qwen_forward_blackbox_stage2.py}.
\end{enumerate}
The base settings are \(n_1=192,n_{\mathrm{out}}=128,m=3072,d=64,\rho=4.0710381,h=0.003\), FP64 output, and seed~0; Stage~1 uses five starts and 15 outer iterations, and Stage~2 uses 12 iterations with a 1,200-iteration conjugate-gradient least-squares (CGLS) cap. This reproduces the 9,437,184-query base observation and Stages~1--2, not the full-input headline rows. Those additionally require Sections~\ref{nested-dimensional-decomposition-scaling-input-dimension-by-stages}--\ref{empirical-evaluation}'s extension, polish, and query-only selection. Checkpoint values enter only the final evaluation.

\bibliographystyle{ACM-Reference-Format}
\bibliography{paper_p1_aisec}

\textbf{Generative AI use disclosure.} OpenAI Codex and Anthropic Claude assisted with implementation, experiment orchestration and analysis, and manuscript editing. The authors inspected generated code, executed and audited the reported experiments, checked numerical results against saved artifacts, and verified citations and claims against source materials. The authors conceived the research questions and attack design and take responsibility for the paper.

\end{document}

%% file: paper_p1_arxiv_metadata.tex
\def\PoneArxiv{}

\def\PoneArxivAuthors{%
  \author{Chunhui Shi}%
  \email{cshi@avitam.ai}%
  \author{Xinwen Fu}%
  \email{xinwen_fu@uml.edu}%
  \renewcommand{\shortauthors}{Shi and Fu}%
}

\def\PoneArxivPdfMetadata{%
  \AtBeginDocument{%
    \hypersetup{%
      pdfauthor={Chunhui Shi and Xinwen Fu},
      pdfkeywords={model extraction, cryptanalysis, neural networks, large language models, GLU, SwiGLU, black-box attacks}%
    }%
  }%
}

%% file: figures/scope.tex
\begin{figure}[t]
\centering
\begin{tikzpicture}[
  font=\scriptsize,
  box/.style={draw, rounded corners=2pt, align=center, minimum height=7mm, minimum width=37mm, inner xsep=4pt},
  target/.style={box, fill=blue!8, very thick},
  outside/.style={box, fill=gray!8},
  stackback/.style={box, fill=gray!14, inner sep=0pt},
  arrow/.style={-{Latex[length=1.8mm]}, thick},
  note/.style={align=left, text width=19mm, font=\scriptsize\itshape}
]
\node[outside] (tok) {API input \(z\)\\{\tiny text $\to$ token ids}};
\node[outside, below=3.5mm of tok] (emb) {embedding layer\\{\tiny initial representation \(r_0=E[\mathrm{ids}]\)}};
\node[outside, below=3.5mm of emb] (attn) {preceding Transformer computations\\{\tiny produce FFN input \(x\)}};
\node[target, below=3.5mm of attn] (ffn) {isolated bias-free GLU FFN\\{\footnotesize $y=f(x)=W_o(\sigma(W_gx)\odot W_ux)$}};
\node[stackback, below=3.5mm of ffn, xshift=1.8mm, yshift=1.8mm] {};
\node[stackback, below=3.5mm of ffn, xshift=0.9mm, yshift=0.9mm] {};
\node[outside, below=3.5mm of ffn] (rest) {residual composition $+$ later blocks\\{\tiny repeated attention $+$ FFN}};
\node[outside, below=3.5mm of rest] (head) {language-model head\\{\tiny $\to$ API-visible output \(F(z)\)}};
\draw[arrow] (tok) -- (emb);
\draw[arrow] (emb) -- (attn);
\draw[arrow] (attn) -- (ffn);
\draw[arrow] (ffn) -- (rest);
\draw[arrow] (rest) -- (head);
\node[note, right=3mm of attn] {outside this experiment};
\node[note, right=3mm of ffn] {block-level interface};
\node[note, right=3mm of rest] {outside this experiment};
\end{tikzpicture}
\caption{Information flow through a complete Transformer model. Preceding computations produce the target FFN input \(x\); the FFN maps \(x\) to \(y=f(x)\); residual composition, later blocks, and the language-model head produce the API-visible output \(F(z)\). Our recovery target is the isolated GLU FFN block.}
\Description{A vertical flow from text input through tokenization, embedding, preceding Transformer computations, an isolated bias-free GLU FFN, later blocks, and the language-model head. Only the GLU FFN is highlighted as the experimental target.}
\label{fig:scope}
\end{figure}
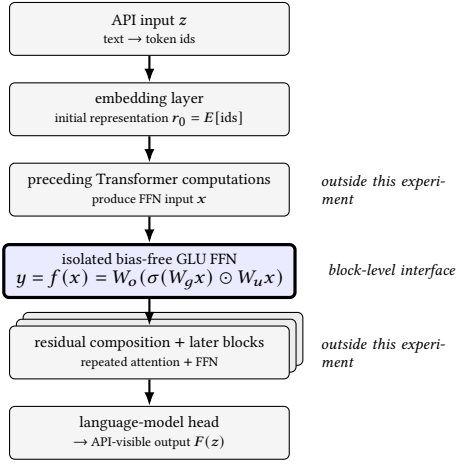

%% file: figures/separation.tex
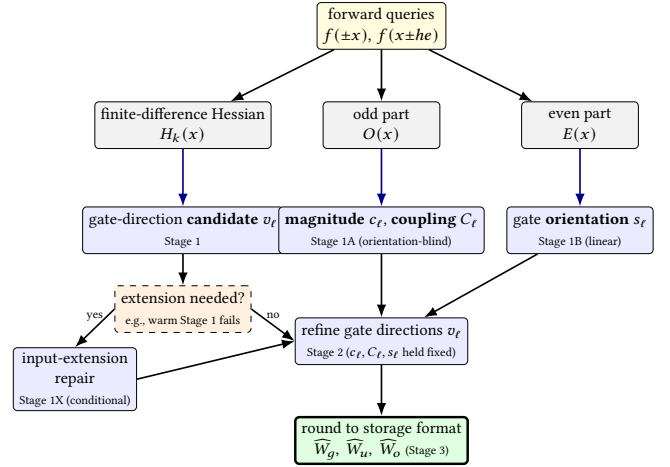
\begin{figure}[t]
\centering
\resizebox{\columnwidth}{!}{%
\begin{tikzpicture}[
  font=\small,
  box/.style={draw, rounded corners=2pt, align=center, minimum height=8mm, minimum width=20mm, inner xsep=3pt, inner ysep=2pt},
  src/.style={box, fill=yellow!15},
  view/.style={box, fill=gray!10},
  stage/.style={box, fill=blue!8},
  cond/.style={box, fill=orange!12, dashed},
  result/.style={box, fill=green!12, very thick},
  arr/.style={-{Latex[length=2.2mm]}, thick},
  sep/.style={-{Latex[length=2.2mm]}, thick, blue!55!black}
]
\node[src] (ora) at (0,4.2) {forward queries\\$f(\pm x),\,f(x{\pm}h e)$};
\node[view] (hess) at (-3.4,2.5) {finite-difference Hessian\\$H_k(x)$};
\node[view] (odd)  at (0,2.5)    {odd part\\$O(x)$};
\node[view] (even) at (3.4,2.5)  {even part\\$E(x)$};
\node[stage] (dir)  at (-3.4,0.7) {gate-direction \textbf{candidate} $v_\ell$\\{\scriptsize Stage 1}};
\node[stage] (mag)  at (0,0.7)    {\textbf{magnitude} $c_\ell$, \textbf{coupling} $C_\ell$\\{\scriptsize Stage 1A (orientation-blind)}};
\node[stage] (sign) at (3.4,0.7)  {gate \textbf{orientation} $s_\ell$\\{\scriptsize Stage 1B (linear)}};
\node[cond]  (needx) at (-3.4,-0.65) {extension needed?\\{\scriptsize e.g., warm Stage 1 fails}};
\node[stage] (xdir)  at (-5.25,-1.85) {input-extension\\repair\\{\scriptsize Stage 1X (conditional)}};
\node[stage]  (ref)  at (0,-1.2) {refine gate directions $v_\ell$\\{\scriptsize Stage 2 ($c_\ell,C_\ell,s_\ell$ held fixed)}};
\node[result] (snap) at (0,-2.9) {round to storage format\\$\widehat W_g,\ \widehat W_u,\ \widehat W_o$\ {\scriptsize (Stage 3)}};
\draw[arr] (ora) -- (hess);
\draw[arr] (ora) -- (odd);
\draw[arr] (ora) -- (even);
\draw[sep] (hess) -- (dir);
\draw[sep] (odd)  -- (mag);
\draw[sep] (even) -- (sign);
\draw[arr] (dir) -- (needx);
\draw[arr] (needx.west) -- node[above, font=\scriptsize] {yes} (xdir.north);
\draw[arr] (xdir.east) -- (ref.west);
\draw[arr] (needx.east) -- node[above, font=\scriptsize] {no} (ref.west);
\draw[arr] (mag)  -- (ref);
\draw[arr] (sign) -- (ref);
\draw[arr] (ref) -- (snap);
\end{tikzpicture}%
}
\caption{\textbf{Separation by observation.} The finite-difference Hessian supplies gate-direction candidates; the odd and even parts of $f(\pm x)$ inform the magnitude/coupling and orientation subproblems. Stage~2 refines the result and Stage~3 rounds to the declared storage format.}
\Description{A flowchart maps finite-difference Hessians to gate-direction candidates, odd outputs to magnitudes and couplings, and even outputs to orientations. The branches merge for direction refinement and storage-format rounding; a dashed conditional branch represents input-extension repair.}
\label{fig:pipeline}
\end{figure}